\documentclass[sigconf]{acmart}
\RequirePackage[shortlabels,inline]{enumitem}
\setlist[itemize]{noitemsep,leftmargin=*,topsep=0em}
\setlist[enumerate]{noitemsep,leftmargin=*,topsep=0em}
\AtBeginDocument{%
  }

\setcopyright{acmlicensed}
\copyrightyear{2025}
\acmYear{2025}
\acmDOI{XXXXXXX.XXXXXXX}
\acmConference[MM ’25]{In Proceedings of the 33rd ACM International Conference on Multimedia}{October 27--31,
  2025}{Dublin, Ireland}

\begin{document}

\title{Tabular Incremental Inference}

\author{Xinda Chen}
\email{chenxd23@m.fudan.edu.cn}
\affiliation{%
  \institution{Fudan University}
  \city{Shanghai}
  \country{China}
}

\author{Zhen Xing}
\email{xingz20@fudan.edu.cn}
\affiliation{%
  \institution{Fudan University}
  \city{Shanghai}
  \country{China}
}

\author{Hanyu Zhang}
\email{hanyuzhang24@m.fudan.edu.cn}
\affiliation{%
  \institution{Fudan University}
  \city{Shanghai}
  \country{China}
}

\author{Weimin Tan*}
\email{wmtan@fudan.edu.cn}
\affiliation{%
  \institution{Fudan University}
  \city{Shanghai}
  \country{China}
}

\author{Bo Yan*}
\email{byan@fudan.edu.cn}
\affiliation{%
  \institution{Fudan University}
  \city{Shanghai}
  \country{China}
}

\renewcommand{\shortauthors}{X Chen et al.}

\begin{abstract}
Tabular data is a fundamental form of data structure. The evolution of table analysis tools reflects humanity's continuous progress in data acquisition, management, and processing. The dynamic changes in table columns arise from technological advancements, changing needs, data integration, etc. However, the standard process of training AI models on tables with fixed columns and then performing inference is not suitable for handling dynamically changed tables. Therefore, new methods are needed for efficiently handling such tables in an unsupervised manner. In this paper, we introduce a new task, Tabular Incremental Inference (TabII), which aims to enable trained models to incorporate new columns during the inference stage, enhancing the practicality of AI models in scenarios where tables are dynamically changed. Furthermore, we demonstrate that this new task can be framed as an optimization problem based on the information bottleneck theory, which emphasizes that the key to an ideal tabular incremental inference approach lies in minimizing mutual information between tabular data and representation while maximizing between representation and task labels. Under this guidance, we design a TabII method with Large Language Model placeholders and Pretrained TabAdapter to provide external knowledge and Incremental Sample Condensation blocks to condense the task-relevant information given by incremental column attributes. Experimental results across eight public datasets show that TabII effectively utilizes incremental attributes, achieving state-of-the-art performance.
\end{abstract}




\maketitle

\section{Introduction}
Tabular data, a foundational structure for organizing information, has evolved alongside humanity's advancements in data collection, management, and analysis \cite{DBLP:journals/corr/abs-2106-03253}. Ubiquitous across fields such as healthcare, climate science, and finance, tabular data has driven the development of diverse algorithms, from gradient-boosted decision trees to neural networks \cite{9551946,mcelfresh2023neuralnetsoutperformboosted,ganin2015unsupervised}. Despite these advancements, existing methods struggle to accommodate incremental columns, which represent new attribute dimensions. Traditional tabular learning models typically assume a fixed column size, which limits their ability to adapt to new information. This inflexibility poses significant challenges, as the inability to incorporate new features can lead to sub-optimal performance and hinder the model's applicability in evolving data environments. 

\begin{figure}[t]
\centering
\includegraphics[width=0.95\columnwidth]{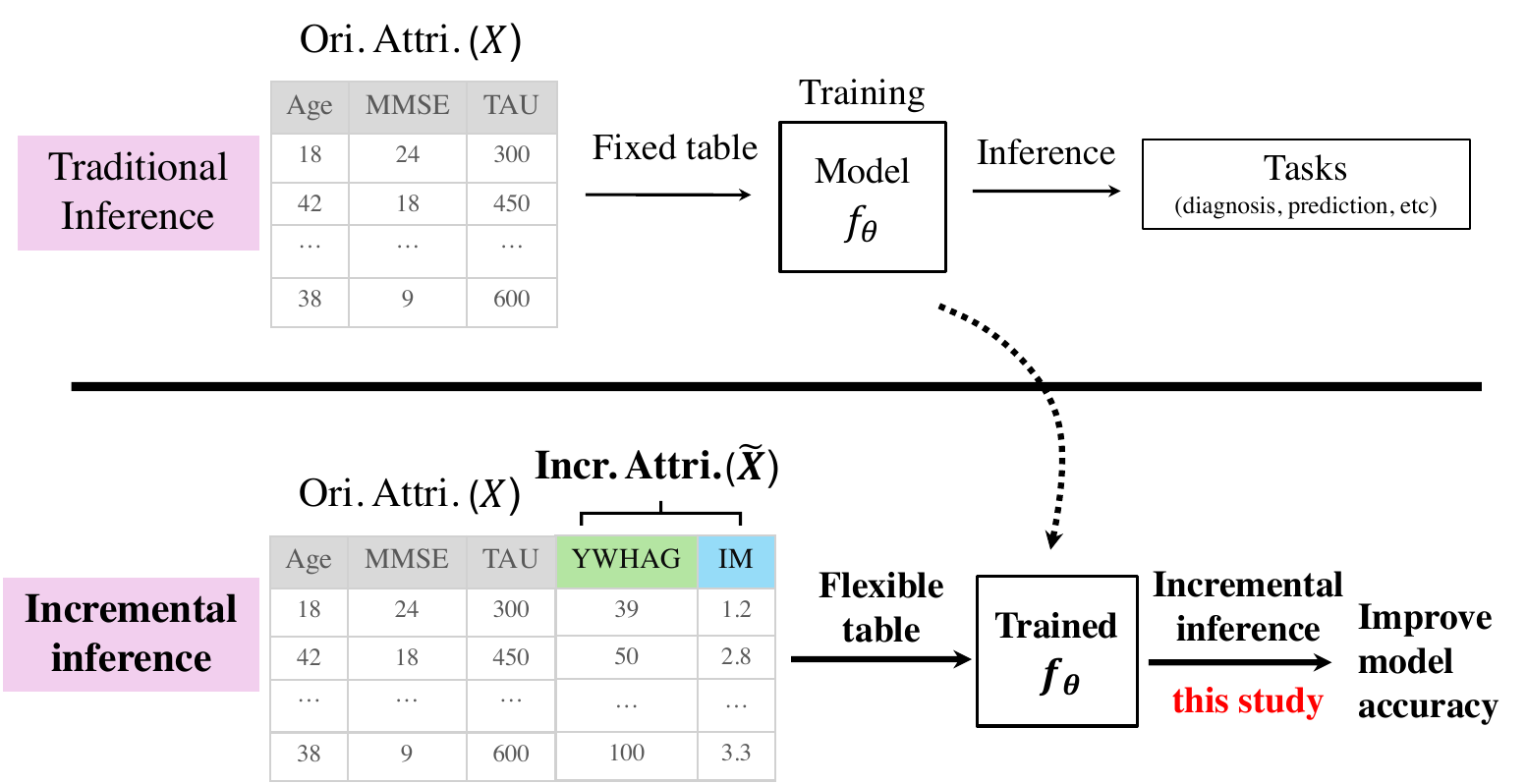}
\caption{Illustration of tabular incremental inference, where model $f_\theta$ trained on original attributes $X$ is adapted during inference to incorporate incremental attributes $\tilde{X}$, like significant factors recently proposed in Alzheimer's disease prediction: YWHAG in Nat. Hum. Behav. \cite{Guo2024} and MI in Nat. Med. \cite{Jack2024}. The goal is to enhance model performance by effectively integrating the information from these new attributes into the trained model.}
\label{fig1}
\end{figure}

This challenge highlights the importance of addressing the task of \textbf{Tabular Incremental Inference}, which involves adapting a trained tabular learning model to incorporate incremental attributes encountered during the inference stage that were not present during training. As shown in Fig. 1, we use Alzheimer’s Disease Prediction as an example. The training data consists of original attributes \( X \), including the patient’s age, Mini-Mental State Examination (MMSE) score, and TAU marker, along with corresponding classification labels, which are used to train the initial model \( f_\theta \). This model is then employed during the inference stage to perform incremental inference on data containing new attributes \( \tilde{X} \), such as YWHAG \cite{Guo2024} and inflammation/immune-related markers (IM) \cite{Jack2024}, which were recently proposed in Nature sub-journals and shown to have a significant impact on Alzheimer’s disease. The task requires the model to adapt to new data attributes, effectively integrate the additional information brought by these incremental attributes, and enhance predictive performance in an unsupervised manner, as it is not feasible to recall patients from the training set for re-evaluation of newly discovered biomarkers. Therefore, this approach is especially beneficial in dynamic environments where data continuously evolves. Models capable of incremental inference can eliminate the need for extensive data preparation and costly retraining when new attributes are introduced, thus improving performance.

However, designing a powerful model presents significant challenges. A key issue is managing the disparity in attribute lengths, as conventional methods cannot accommodate new attributes. Traditional machine learning models and neural networks rely on fixed input sizes, leading them to ignore incremental attributes during inference, which can degrade performance. Effectively utilizing these incremental columns is another challenge, particularly since tabular data is heterogeneous, with continuous, categorical, and ordinal values that vary widely \cite{park2022hrgan,nam2023stuntfewshottabularlearning}. Directing applying the original model to incremental attributes can decrease accuracy, contrary to the expectation that additional dimensions should enhance performance. Existing methods, whether supervised, self-supervised, or assisted by Large Language Models (LLMs), are constrained by these issues, struggling to integrate incremental attributes without undermining the structural strengths of tabular data.

We analyze this task through the lens of Information Bottleneck (IB) theory, proving that putting incremental attributes into usage will improve IB. The key for a model to benefit from incremental attributes lies in maximizing the retention of task-relevant information while filtering out irrelevant data, which can be quantified by mutual information. A detailed proof is provided in the method section. To achieve the optimization of IB, we introduce TabII with Placeholders and Incremental Sample Condensation (ISC) blocks. Placeholders leverage external knowledge, including the name and meaning brought by LLM as well as understanding of tabular structure and correlation given by Tabular Foundation Models like TabPFN v2, thereby enriching the information available for incremental attributes and helping the model better grasp relevant information. ISC blocks guide the adaptation of incremental attributes by specially designed attention mechanisms, benefiting both the extraction of relevant information and the discard of irrelevant. Furthermore, through mutual information estimation methods, we quantitatively demonstrate that TabII significantly increases the mutual information between learned representations and labels. In summary, the contributions of this paper are:
\setlist[itemize]{leftmargin=*}

\begin{itemize}
  \setlength\itemsep{0em} 
  \item This paper introduces a new task, Tabular Incremental Inference, aiming to enable trained models to incorporate new columns during the inference stage, enhancing the practicality of AI models in dynamically changed tables.
  \item We present a description of this new task within the framework of information bottleneck theory. Maximizing mutual information between task labels and learned representations while minimizing between representations and tabular data is crucial for the model to effectively leverage incremental attributes during the inference stage.
  \item Following the information bottleneck theory, we propose a TabII method with Placeholders and Incremental Sample Condensation blocks for better incremental learning ability. Our approach achieves state-of-the-art performance across eight public datasets. Additionally, we quantify mutual information to validate model efficacy.
\end{itemize}

\section{Related Work}
\subsection{Tabular Contrastive Learning}
Contrastive learning learns meaningful representations without labeled data, making it effective in pretraining. Its core idea is to model similarity between augmented views of the same sample and dissimilarity between different samples \cite{bahri2021scarf}. Various strategies optimize contrastive loss by defining sample pairs through data augmentations like masking or cropping in feature space \cite{ucar2021subtab,he2020momentumcontrastunsupervisedvisual}. These techniques capture underlying data structures, leading to robust representations. Recent work extends contrastive learning to the tabular domain; for example, SCARF \cite{bahri2021scarf} uses random masking on sample columns to construct positive pairs, while TransTab \cite{wang2022transtablearningtransferabletabular} introduces Vertical-Partition Contrastive Learning by partitioning tabular columns to create positive and negative samples.

However, neither method is ideal for tabular incremental inference. SCARF’s requirement for fixed input sizes limits its flexibility. TransTab, while achieving robust table representations, may sacrifice downstream task performance by overlooking the incremental relationships between existing and new data. Research specifically addressing contrastive learning in Tabular Incremental Inference remains sparse. Our proposed TabII method addresses this gap.

\subsection{LLM Assisted Tabular Learning}
The impressive achievements of Large Language Models (LLMs), scaled to unprecedented sizes and trained on extensive text corpora, have demonstrated their vast knowledge and exceptional capabilities in transfer learning \cite{brown2020language,ouyang2022training}. This success has spurred interest in applying LLMs to tabular deep learning, aiming to harness their extensive world knowledge. Pioneering studies have explored approaches such as fine-tuning LLMs on tabular datasets and using them to generate supplementary or synthetic tabular data \cite{dinh2022lift,fang2024largelanguagemodelsllmstabular}. Some methods tokenize column names and values for LLM input \cite{wang2022transtablearningtransferabletabular}, while others use textual descriptions of columns to enrich the data \cite{hegselmann2023tabllm,zhang2024towards,carballo2023tabtextflexiblecontextualapproach}.

However, these approaches focus on few-shot learning, sacrificing supervised performance and the structured strengths of tabular data. LLMs struggle with small numerical differences, hindering effective integration with traditional tabular models \cite{yan2024makingpretrainedlanguagemodels,meet}. Our task requires a model that adapts quickly to new information while preserving structural knowledge. Thus, it is crucial to develop approaches that combine the structural and numerical strengths of tabular data with the rich contextual knowledge of LLMs to enhance performance. The recent emergence of tabular foundation models, such as TabPFN v2 \cite{Hollmann2025}, represents a significant advancement, enabling the model to understand dynamic inputs. Fine-tuning pre-trained models to incorporate world knowledge presents a promising direction, and this strategy is a central element of the method proposed in this paper.
\section{Method}
In this section, we first give a clear mathematical definition of Tabular Incremental Inference and illustrate this task under information bottleneck theory. We prove that leveraging incremental attributes will improve information bottleneck. Inspired by this, we developed the TabII method that can efficiently utilize incremental attributes in reference.

\subsection{Problem Formulation}

Consider a training tabular set $\mathcal{D}_{\text{train}} = \{(x_i, y_i)\}_{i=1}^N$, where $x_i \in \mathbb{R}^d$ are the input columns and $y_i \in \mathbb{R}$ are the corresponding labels. We train a model $\mathcal{M}$ to learn the mapping function $f_{\theta}: \mathbb{R}^d \rightarrow \mathbb{R}$, parameterized by $\theta$, which predicts the labels from the input columns. During inference, we encounter incremental columns $\tilde{x}_i \in \mathbb{R}^{\tilde{d}}$ that were not available during the training phase. Let $x_i'$ denote the concatenation of the original and incremental columns, such that $x_i' = [x_i; \tilde{x}_i] \in \mathbb{R}^{d + \tilde{d}}$. Our objective is to develop a method that can utilize these incremental columns to improve the prediction accuracy.

We model the probability distribution of the label given the columns as $p(y | x, \tilde{x})$ and aim to approximate this distribution using a model trained on $p(y | x)$. Specifically, we define a new model $\mathcal{M}'$ parameterized by $\theta'$, which takes both the original and incremental attributes as inputs. The goal is to learn the function $f_{\theta'}: \mathbb{R}^{d + \tilde{d}} \rightarrow \mathbb{R}$ that maximizes the likelihood of the observed labels:
\[
\theta' = \arg \max_{\theta'} \sum_{i=1}^N \log p(y_i | x_i, \tilde{x}_i; \theta') \tag{1}
\]
\subsection{Information Bottleneck Analysis}

However, when we encounter incremental columns during inference that were not present in the training set, directly maximizing the likelihood (Eq.1) can be intractable since we do not have the ground truth label of the inference data. Instead, using the Information Bottleneck Method allows us to design a model that utilizes the incremental attributes and effectively balances the retention of useful information and the compression of irrelevant information, thereby improving prediction accuracy \cite{tishby2015deep}.

The Information Bottleneck theory offers an information-theoretic framework to extract the essential information that an input random variable $X \in \mathcal{X}$ holds about an output random variable $Y \in \mathcal{Y}$. Assuming their joint distribution $p(x, y)$, the core information is defined by the mutual information $I(X; Y)$, highlighting the dependence between $X$ and $Y$. In this setup, $Y$ inherently identifies which features of $X$ are relevant or irrelevant. An optimal representation of $X$ would capture these relevant features while compressing $X$ to exclude irrelevant parts that do not aid in predicting $Y$ \cite{hjelm2019learningdeeprepresentationsmutual,kawaguchi2023doesinformationbottleneckhelp}.

\begin{figure}[t]
\centering
\includegraphics[width=1\columnwidth]{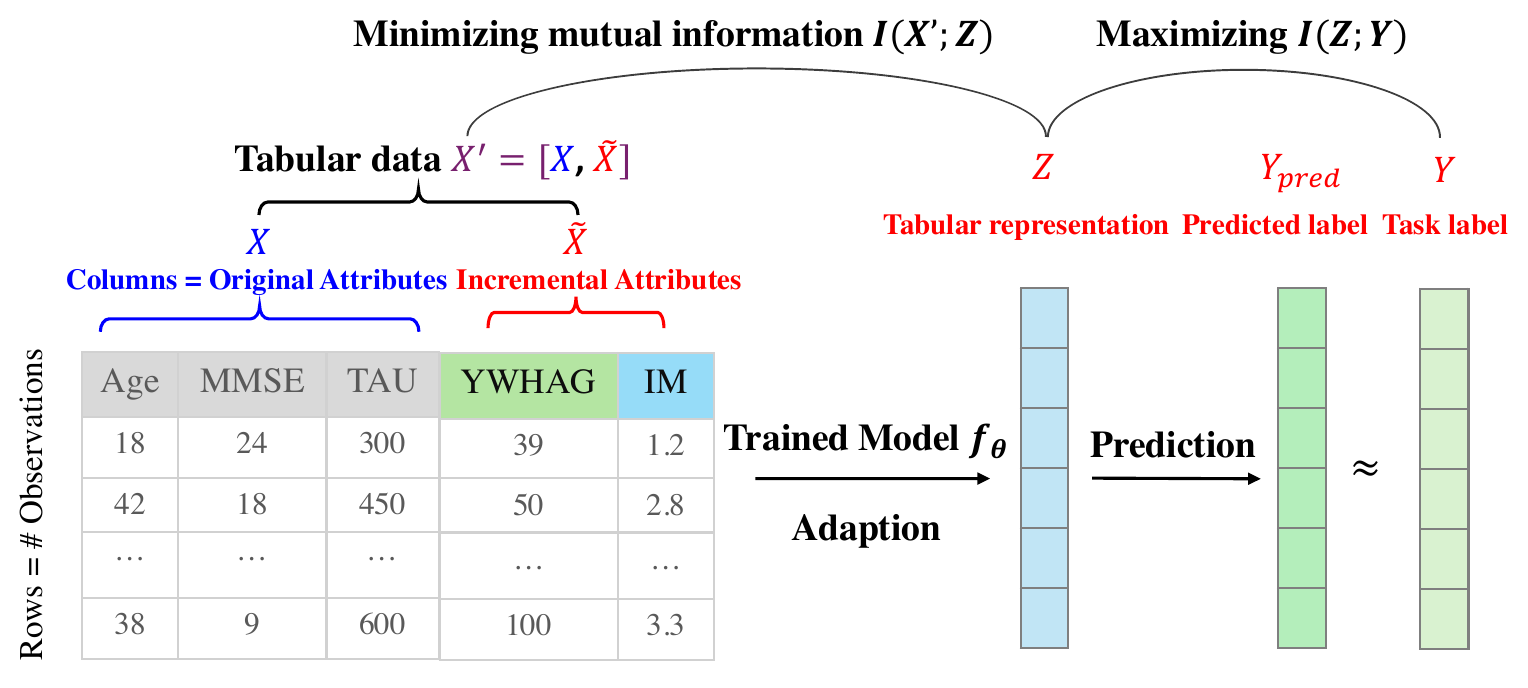}
\caption{Illustration of Incremental Inference guided by Information Bottleneck theory. The task is reframed as an optimization problem: maximizing mutual information between the tabular representation $Z$ from our adaptation model and the task label $Y$ while minimizing mutual information between the tabular data $X'$ and the tabular representation $Z$.}
\label{fig2}
\end{figure}

\begin{figure*}[t]
\centering
\includegraphics[width=1.9\columnwidth]{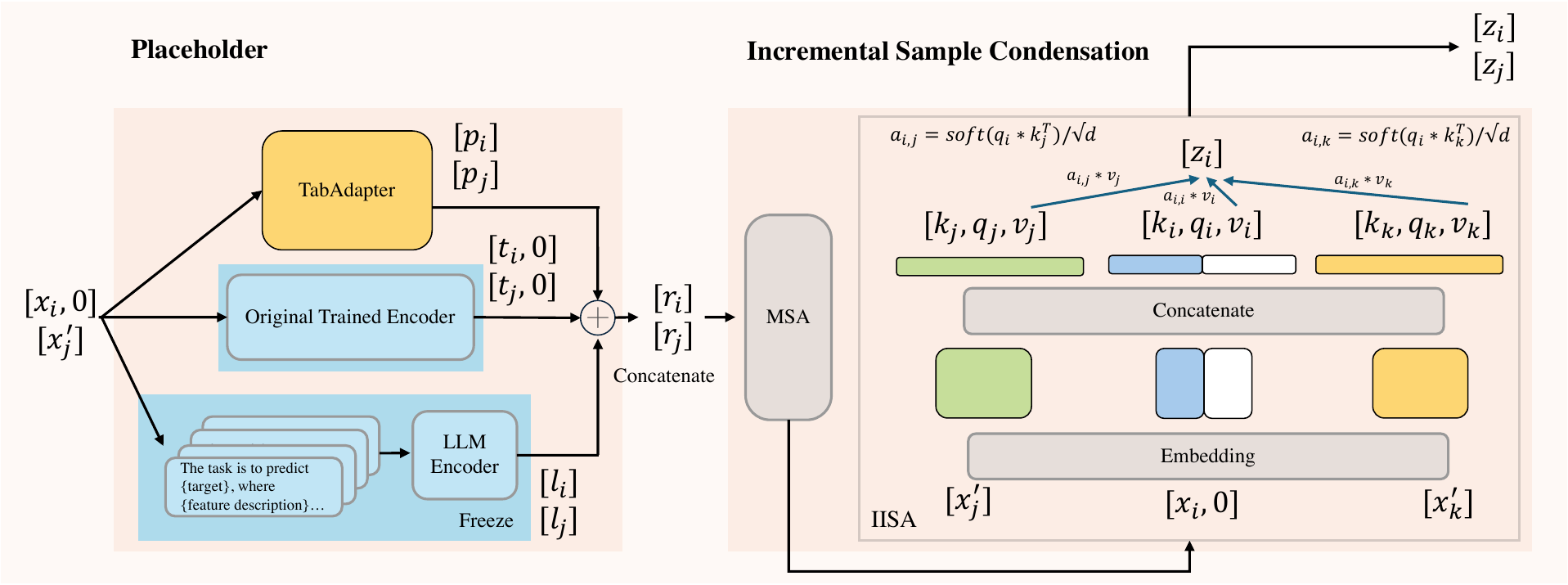}
\caption{Overview of TabII architecture, which consists of two main components inspired by information bottleneck theory to increase $I(Z; Y)$ and decrease $I(X; Z)$: The Placeholder processes the input tabular data, including training samples with appended zeros $[x_i, 0]$ and test samples $x_j'$, splitting them into three pathways and concatenating the outputs from the tabular encoder ($t_i$, $t_j$), TabAdapter ($p_i$, $p_j$) and Large Language Model encoder ($l_i$, $l_j$) to $r_i$ and $r_j$. Incremental Sample Condensation blocks then fuse these modality representations using Multi-head Self-Attention (MSA) and Interior Incremental Sample Attention (IISA) to generate $z_i$ and $z_j$ useful for downstream tasks. $q_i$, $k_i$, and $v_i$ in IISA represent the query, key, and value vectors for the samples, and $d$ is the dimensionality of the key vectors.}
\label{fig3}
\end{figure*}

Statistically, the part of $X$ that is pertinent to $Y$, denoted by $Z$, serves as a minimal sufficient statistic of $X$ for $Y$. This is essentially the simplest transformation of $X$ that retains the mutual information $I(X; Y)$. Under the assumption of the Markov chain $Y \rightarrow X \rightarrow Z$, we aim to minimize the mutual information $I(X; Z)$ to achieve the simplest statistic while maximizing the mutual information $I(Z; Y)$ to improve performance. As shown in Fig. 2, in our task scenario, finding an optimal representation $Z \in Z$ is framed as the maximization of the following function:

\[
\max_{Z} I(Z; Y) - \beta I(Z; X') \tag{2}
\]
where $X'$ is the concatenation of the original and additional input columns, and $\beta$ is a positive Lagrange multiplier that balances the trade-off between the complexity (rate) of the representation, $R = I(X'; Z)$, and the amount of retained relevant information, $I_Y = I(Z; Y)$. We first explore \( I(X'; Z) \) \cite{thomas2006elements}:

\[
I(X'; Z) = H(Z) - H(Z \mid X') \tag{3}
\]
While \( H(Z \mid X') \) remains low and relatively unchanged due to the deterministic nature of the neural network, the entropy \( H(Z) \) measures the uncertainty in \( Z \) itself.
If the original model has not been retrained or adapted and we straightly input a different input distribution into the model, like shuffling the order of attributes or adding incremental attributes in our cases, then \( H(Z) \) is expected to be extremely high. The specific performance of FT-Trans* supports this argument, as depicted in Fig. 4. However, if the model follows a standard SGD optimization, \( I(X'; Z) \) is implicitly minimized and thus varies little \cite{shwartzziv2017openingblackboxdeep}.

Thus we should focus more on $I(Z; Y)$ and we split it similarly \cite{thomas2006elements}:

\[
I(Z; Y) = H(Y) - H(Y\mid Z) \tag{4}
\]
While \( H(Y) \) remains the same for all cases, the conditional entropy \( H(Y\mid Z) \) measures the uncertainty in \( Y \) given \( Z \). Replace \( Z \) by \( f(x) \) in the original model, we have 

\[
H(Y | f(x)) \geq H(Y | f(x), \tilde{x}) \tag{5}
\]
We can prove Eq.5 using the chain rule:
\[
H(Y | f(x)) = I(Y, \tilde{x} | f(x)) + H(Y | f(x), \tilde{x})\tag{6}
\]
Since mutual information \( I(Y, \tilde{x}|f(x))\) is always non-negative. Therefore, Eq.5 says that $I(Z; Y)$ is expected to increase as incremental attributes are utilized, and we need to develop an adaption model to combine the information in the original model and $\tilde{x}$. The next section details how we designed our model to realize a smaller $I(Z; X')$ and larger $I(Z; Y)$. Furthermore, to give a quantitative explanation of how well our model captures appropriate condensation, we use the Mutual Information Neural Estimation (MINE) method to estimate two variables' mutual information \cite{belghazi2018mutual}. MINE maximizes a lower bound on mutual information via a neural network-based objective function. The main formula involves the Donsker-Varadhan representation \cite{Donsker1974}:

\[
I(X; Y) \geq \mathbb{E}_{P_{XY}}[T(x, y)] - \log(\mathbb{E}_{P_X \otimes P_Y}[e^{T(x, y)}]) \tag{7}
\]
where \( T(x, y) \) is a neural network. By training this neural network to maximize the above objective, MINE provides a scalable and flexible way to estimate mutual information. Implementing MINE, We show that our model achieved higher \( I(Z; Y)\) and smaller \( I(X'; Z)\) in Fig. 4.

\subsection{TabII}
Fig. 3 illustrates TabII, a versatile method designed for feature-embedded models to address Tabular Incremental Inference. Feature-embedded models mean the models treat each tabular attribute with an embedded dimension. TabII first allocates placeholders for unseen column attributes during training by utilizing the Large Language Model and pretrained tabular foundation models to generate supplementary information, ensuring that the condensation learned during training effectively transfers to the inference stage. Then, Incremental Sample Condensation (ISC) is used for fusion to improve the adaptability of condensation blocks from the inference data distribution. Finally, our pretraining and fine-tuning strategies are detailed.

\subsubsection{Placeholders: LLM Encoder}

TabII first appends zero placeholders for incremental columns that are absent in the train set $x$. Then for each instance, we generate a textual prompt that contextualizes the tabular data, such as:

\begin{quote}
"The task is to predict {target} using the given tabular data, where {feature description} is initially provided. Later, {incremental features} will be added. The model should pay particular attention to these newly introduced features for improved performance."
\end{quote}
Here, {target}, {feature description}, and {incremental features} are dynamically filled placeholders. We then utilize a pre-trained LLM to encode each textual prompt into a dense vector representation. We use $\mathbf{l}_i$ to denote the text representation of the prompt, 

\subsubsection{Placeholders: TabAdapter}

In addition to the information carried by the column names from the language model, the recent development of tabular foundation models allows us to use pretrained base model parameters, which capture external knowledge related to the table structure. For this, we employ the TabPFN v2 model as our foundational model and fine-tune it using LoRA (Low-Rank Adaptation). LoRA enables efficient fine-tuning by introducing low-rank adaptations to the model's weights. Specifically, we decompose the weight matrix $\mathbf{W}$ into a sum of two low-rank matrices:

\[
\mathbf{W} \approx \mathbf{W}_0 + \Delta \mathbf{W} = \mathbf{W}_0 + A \cdot B^T \tag{8}
\]
where $\mathbf{W}_0$ represents the pretrained weight matrix, $A$ and $B$ are the low-rank matrices, and $\Delta \mathbf{W}$ is the low-rank adaptation introduced during fine-tuning.

Additionally, we incorporate Elastic Weight Consolidation (EWC) loss to prevent the model from deviating too far from the pretrained weights. The EWC loss term is defined as:

\[
\mathcal{L}_{\text{EWC}} = \sum_i \frac{\lambda}{2} F_i (\theta_i - \theta^*_{i})^2 \tag{9}
\]

where $\theta_i$ represents the model parameters after fine-tuning, $\theta^*_{i}$ are the pretrained parameters, $F_i$ are the diagonal elements of the Fisher information matrix, and $\lambda$ is a regularization strength hyperparameter. The EWC loss encourages the model to retain important knowledge from the pretrained parameters while still adapting to new data.

The output of TabAdaptor is denoted as $\mathbf{p}_i$. For instance $i$. To create a comprehensive representation for each instance, we concatenate $\mathbf{l}_i$, $\mathbf{p}_i$ with the original tabular data representation $\mathbf{t}_i$. The concatenated vector $\mathbf{r}_i$ is then passed through ISC blocks for better condensation.

From the view of information bottleneck theory, Placeholders will increase both \( I(X'; Z) \) and \( I(Z; Y) \) compared to the original model or zero-padding method. Specific mathematic induction is shown in appendix Eq.14 and 15. We assume that Placeholders will lead to a more significant influence on \( I(Z; Y) \) rather than \( I(X'; Z) \). Since we want to leverage the external world knowledge, then the information gained on \( I(X'; Z) \) is unavoidable. Experiments in Section 4.3 support our assumption, as depicted in Fig. 4's FT-Trans column and LLM column.

\subsubsection{Incremental Sample Condensation}

Incremental Sample Condensation consists of two key components: the first is multi-head self-attention (MSA), which helps the model better capture the importance and relationships across original and incremental columns. This helps reduce redundant information $I(X; Z)$ and increase $I(Z; Y)$ by reweighting the input columns.

Following MSA is Interior Incremental Sample Attention (IISA), designed to enable the condensation block to learn from the distribution of incremental samples. IISA functions as a type of row attention, where attention is computed across different data points (rows of tabular data) within a batch rather than just across attributes of a single data point. Specifically, we concatenate the embedding of each feature for a single data point and then compute attention across the samples. This approach enhances the representation of a given data point by considering the context provided by other points \cite{SAINT}. Attention is computed exclusively using inference set data to reduce $I(\tilde{X}; Z)$ by discarding irrelevant or noisy aspects while enhancing $I(Z; Y)$ by keeping the most informative parts of the incremental data. The assumption is validated in section 4.3, as demonstrated by the comparison between the LLM column and the LLM\&ISC column in Fig. 4. The attention mechanism is defined as follows:

\[
z_i = \sum_{j \in \{i, \text{inference samples}\}} \text{softmax}\left(\frac{q_i \cdot k_j^T}{\sqrt{d}}\right) v_j \tag{10}
\]
Here, $q_i$, $k_j$, and $v_j$ represent the query, key, and value vectors for the samples, and $d$ is the dimensionality of the key vectors. This final representation $\mathbf{z}_i$ is then fed into the downstream neural network for prediction tasks.

\subsubsection{Adaption}
Our model is adapted through a unified training stage combining contrastive learning and supervised learning with pseudo labels. For each input sample $\mathbf{x} \in \mathbb{R}^{d + \tilde{d}}$, we create two views: the original input $\mathbf{x}$ and a masked version $\tilde{\mathbf{x}}$ where random features are set to zero. The contrastive loss maximizes agreement between these views:

\begin{equation}
\mathcal{L}_{\text{contrastive}} = -\log \frac{\exp(\text{sim}(\mathbf{h}, \tilde{\mathbf{h}})/\tau)}{\sum_{i=1}^N \exp(\text{sim}(\mathbf{h}, \tilde{\mathbf{h}}_i)/\tau)} \tag{11}
\end{equation}
where $\mathbf{h} = f_{\theta'}(\mathbf{x})$, $\tilde{\mathbf{h}} = f_{\theta'}(\tilde{\mathbf{x}})$, $\text{sim}$ is cosine similarity, and $\tau$ is a temperature parameter.

The cross-entropy loss combines labeled training data and pseudo-labeled inference data:

\begin{equation}
\mathcal{L}_{\text{CE}} = \frac{1}{|\mathcal{D}_{\text{train}}|} \sum_{\mathcal{D}_{\text{train}}} \mathcal{L}(f_{\theta'}(\mathbf{x}_i), y_i) + \frac{\lambda}{|\mathcal{D}_{\text{pseudo}}|} \sum_{\mathcal{D}_{\text{pseudo}}} \mathcal{L}(f_{\theta'}(\mathbf{x'}_i), \tilde{y}_i) \tag{12}
\end{equation}
where $\tilde{y}_i$ are pseudo-labels generated by the original model $f_\theta$, and $\lambda$ balances the pseudo-label contribution.

The total adaptation loss is:

\begin{equation}
\mathcal{L}_{\text{total}} = \mathcal{L}_{\text{CE}} + \alpha\mathcal{L}_{\text{contrastive}} + \mathcal{L}_{\text{EWC}} \tag{13}
\end{equation}
with $\alpha$ controlling contrastive learning importance and $\mathcal{L}_{\text{EWC}}$ defined previously.

\begin{figure}[t]
\centering
\includegraphics[width=1\columnwidth]{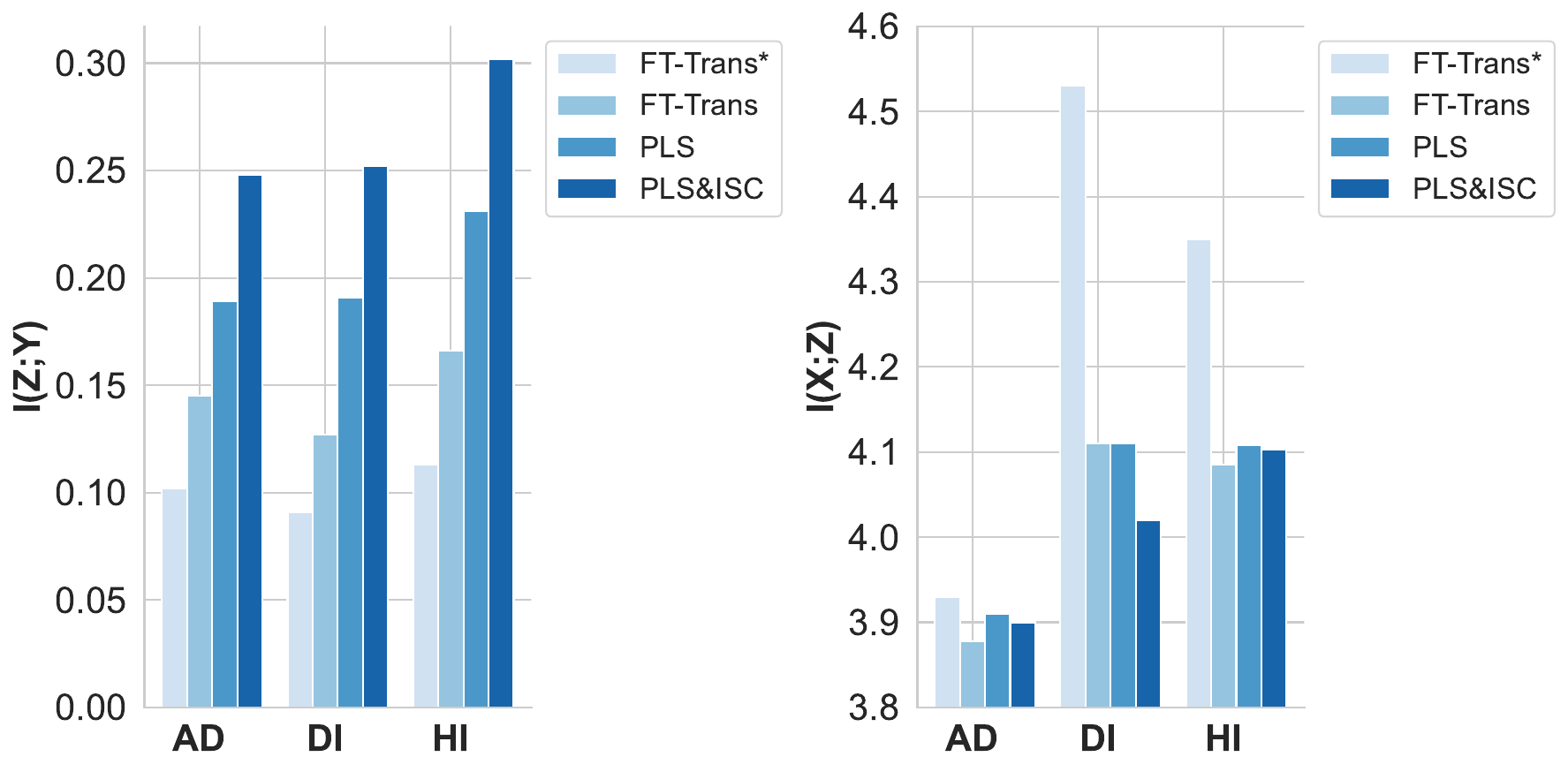}
\caption{Estimation of mutual information using FT-Trans*, FT-Trans, TabII's Placeholder, and full TabII on three datasets. The left side shows $I(Z; Y)$ while the right shows $I(X; Z)$; results show that TabII generally achieved the largest $I(Z; Y)$ and smaller $I(X; Z)$, representing great incremental learning ability.}
\label{fig6}
\end{figure}

\begin{table*}[ht]
\centering
\begin{tabular}{lcccccccc|c}
\toprule
\textbf{Dataset size} & 768 & 48842 & 500 & 32561 & 1000 & 98050 & 7043 & 18472\\
\textbf{Train attribute size} & 4 & 4 & 4 & 8 & 3 & 5 & 9 & 25\\
\textbf{Incremental attribute size} & 4 & 10 & 8 & 6 & 17 & 23 & 10 & 60\\
\midrule
\textbf{Method/Dataset} & \textbf{DI} & \textbf{AD} & \textbf{DS} & \textbf{IC} & \textbf{CG}  & \textbf{HI} & \textbf{BL} & \textbf{IO} & \textbf{\textbf{Rank(Std)}}\\
\midrule
LR & 0.680 & 0.818 & 0.572 & 0.840 & 0.670 & 0.547 & 0.781 & 0.637 & 9.88(3.02)\\
XGBoost & 0.646 & 0.817 & 0.508 & 0.855 & 0.682 & 0.556 & 0.768 & 0.785 & 9.75(5.04)\\
CatBoost & 0.656 & 0.817 & 0.572 & 0.855 & 0.674 & 0.564 & 0.774 & 0.790 & 7.44(3.94)\\
MLP & 0.669 & 0.817 & 0.580 & 0.841 & 0.672 & 0.563 & 0.782 & 0.773 & 8.06(2.69)\\
ResNet & 0.685 & 0.818 & 0.546 & 0.842 & 0.679 & 0.563 & 0.786 & 0.786 & 6.75(3.28)\\
TabNet & 0.668 & 0.815 & 0.562 & 0.837 & 0.660 & 0.560 & 0.768 & 0.705 & 11.75(1.67)\\
SCARF & 0.672 & 0.818 & 0.560 & 0.844 & 0.672 & 0.560 & 0.782 & 0.744 & 8.56(2.76)\\
FT-Trans & 0.698 & 0.819 & 0.580 & 0.843 & 0.686 & 0.563 & 0.784 & 0.722 & 5.69(2.25)\\
\midrule
FT-Trans* & 0.653 & 0.762 & 0.572 & 0.747 & 0.670 & 0.537 & 0.732 & 0.550 & 14.56(2.80)\\ 
TransTab & 0.713 & 0.812 & 0.524 & 0.837 & 0.698 & 0.626 & 0.785 & 0.637 &8.25(5.44)\\
\midrule
TabLLM & 0.680 & 0.804 & 0.530 & 0.840 & 0.530 & 0.540 & 0.762 & 0.628 & 13.31(2.34)\\
MediTab & 0.688 & 0.814 & 0.540 & 0.854 & 0.684 & 0.544 & 0.758 & 0.622 & 10.75(4.40)\\
\midrule
TabPFN-v2 ICL with train set only & 0.680 & 0.802 & 0.581 & 0.836 & 0.692 & 0.623 & 0.764 & 0.719 & 9.38(4.37) \\
TabPFN-v2 ICL add test without label & 0.695 & 0.815 & 0.605 & 0.845 & 0.706 & 0.647 & 0.772 & 0.745 & 5.69(3.07) \\
TabPFN-v2 ICL add test with pseudo label & 0.695 & 0.822 & 0.608 & 0.832 & 0.718 & 0.672 & 0.781 & 0.745 & 5.19(4.54)\\
\midrule
\textbf{TabII} & \textbf{0.749} & \textbf{0.854} & \textbf{0.690} & \textbf{0.892} & \textbf{0.752} & \textbf{0.732} & \textbf{0.826} & \textbf{0.802} & \textbf{1.00(0.00)}\\
\bottomrule
\end{tabular}%
\caption{Comparison of different methods on public datasets. The evaluation index is accuracy. For each dataset, the best results are shown in bold. Reported results are averaged over four trials. The rank column reports the average rank across all datasets.}
\label{tab:results}
\end{table*}

\section{Experiments}
\subsection{Experimental Setup}
We use eight public datasets that are commonly used in tabular area: Diabetes (DI) \cite{Diabetes}, Adult (AD) \cite{misc_adult_2}, Dress-sales (DS) \cite{misc_dresses_attribute_sales_289}, Income (IC) \cite{Income}, Credit-g (CG) \cite{misc_statlog_(german_credit_data)_144}, Higgs (HI) \cite{misc_higgs_280}, Blastchar (BL) \cite{Blastchar}, and Insurance-co (IO) \cite{misc_insurance_company_benchmark_(coil_2000)_125}. In each, we remove specific columns from the train and validation sets while keeping them in the inference set to simulate real-world incremental tabular inference. The deleted columns were chosen logically, reflecting their historical appearance. Details are in the appendix. All algorithms use the same train-validation-test split of 6:2:2, with the test set further divided into TrainFromTest, ValFromTest, and TestFromTest. TrainFromTest and ValFromTest without labels are used in the training stage in TabII, while TestFromTest is reserved for inference to avoid information leakage.

For baselines, we use logistic regression, XGBoost \cite{xgboost}, and CatBoost \cite{catboost}, deep learning models like MLPs, ResNet \cite{resnet}, TabNet \cite{tabnet}, SCARF \cite{bahri2021scarf}, and FT-Transformer \cite{FT-transformer}. Since these methods can’t utilize incremental columns, we only use the same columns as the train set for inference. Also, we compare our method to FT-Trans* and TransTab \cite{wang2022transtablearningtransferabletabular}, which can utilize all columns in the inference set. In our experiments, we modified FT-Transformer by replacing the embedding layer with a linear embedding to handle any input dimensions, noted as FT-Trans*. Finally, for fairness, we compare performance with LLM-assisted models, which also use external knowledge. They include TabLLM \cite{hegselmann2023tabllm} and MediTab \cite{wang2024meditabscalingmedicaltabular}. Finally, we compare with TabPFN v2 \cite{Hollmann2025}'s zero-shot performance, in-context learning without labels as well as in context learning with pseudo labels as provided in section 3. We omit comparison with methods that generate missing columns from original ones, as information bottleneck theory suggests no information gain in such generation, making performance equivalent to discarding those columns. Data preprocess and implementation details are also shown in the appendix.

\begin{figure}[t]
\centering
\includegraphics[width=0.61\columnwidth]{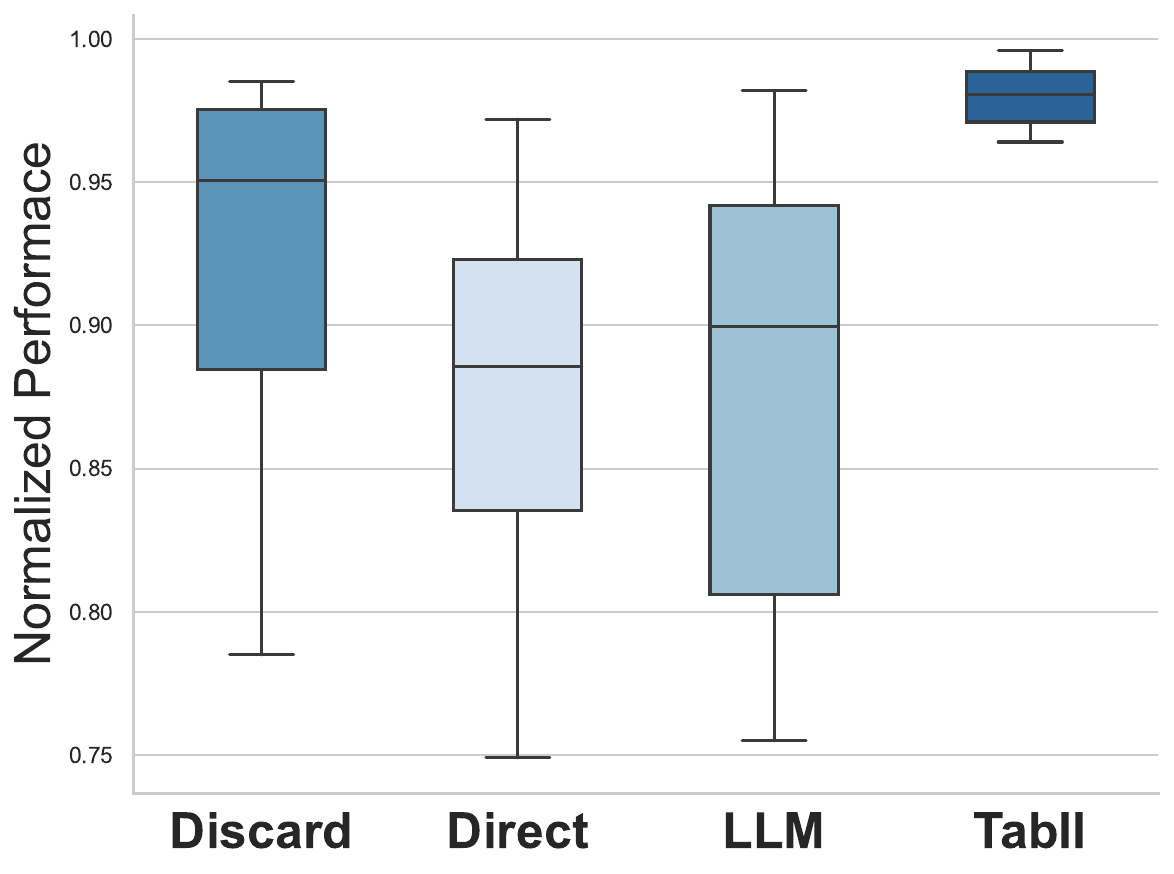}
\caption{Box plot showing the average comparative performance across eight datasets for each method group, which equals the method accuracy divided by the optimal accuracy where the model is trained on fully supervised datasets with incremental attributes included. Discard represents methods that neglect incremental attributes. Direct refers to using the original model for incremental attributes. The figure shows that TabII consistently delivers excellent performance, averaging 97\% of the optimal situation.}
\label{fig4}
\end{figure}

\subsection{Results on Public Datasets}
We conducted a comprehensive performance comparison of TabII against existing methods across eight public datasets, as presented in Table 1. The analysis reveals that TabII consistently outperforms other approaches by effectively leveraging the information provided by incremental attributes during inference. Also, TabII converges five times faster than retraining an FT-Trans if labels are available. Each trial is repeated four times to avoid randomness, and the variance of our method is shown in the appendix. To present this observation more intuitively, we grouped the methods from LR to FT-Trans into the Discard group since these methods cannot utilize incremental attributes and rely solely on the original model. The Direct group refers to FT-Trans*, where the incremental attributes are directly input into the originally trained model without any adaptation. The LLM group includes TabLLM and MediTab which also use large language models' knowledge for reference. We normalized the accuracy based on the optimal scenario where the model is trained on a fully supervised dataset with incremental attributes included. In other words, Fig. 5 shows the comparative performance relative to each dataset's optimal scenario. The box plot illustrates the performance of each method group. The Discard group shows an average performance decrease of 5\%, while the Direct group experiences a more significant decline, indicating that the original model fails to capture the information from incremental attributes effectively. The LLM-assist group also performs poorly since they ignore the numerical structure and the relationship between original and incremental attributes. In contrast, TabII boosts the model's performance above the Discard group, achieving 97\% of the ideal performance on average. Finally, we compare TabII with models also using pretrained TabPFN v2's weights. While doing in-context learning with both train and inference data with pesudo labels using TabPFN v2 achieving best score among its family. Both the accuracy and stability are much worse than TabII.

\begin{figure}[t]
\centering
\includegraphics[width=0.92\columnwidth]{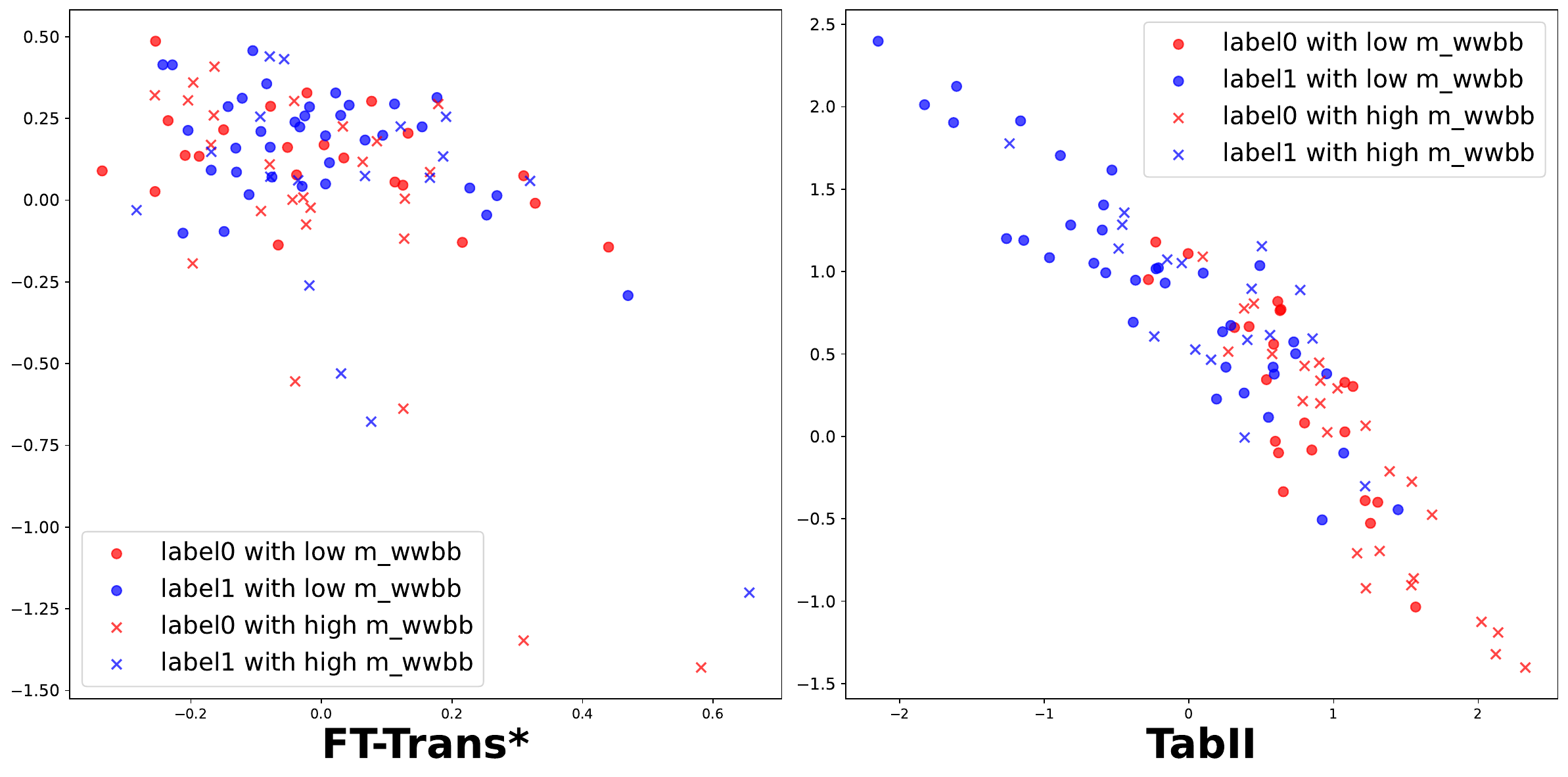}
\caption{Representation visualization after PCA on HI dataset through FT-Trans* and TabII. The color indicates task labels, while the marker denotes an important incremental attribute m\_wwbb. The results show that FT-Trans* fails to capture the information in m\_wwbb while TabII changes representation locations according to m\_wwbb value for better separation of classes.}
\label{fig5}
\end{figure}

We also present visualization results of TabII and FT-Trans* on the HI dataset in Fig. 6 to illustrate TabII's effective adaptation and learning from incremental attributes while FT-Trans* fails. Data points were scatter plotted after dimension reduction by PCA \cite{MACKIEWICZ1993303}. Original attributes with one incremental attribute m\_wwbb are input for reference. FT-Trans cannot use the information contained in m\_wwbb, while TabII changes the representation according to m\_wwbb value, leading to better label separation.

\subsection{Mutual Information Estimation}
We evaluate the effectiveness of  TabII's Placeholder part and ISC blocks by estimating mutual information between layers using MINE. Specifically, we estimate $I(Z; Y)$ and $I(X; Z)$ for TabII with Placeholder only, TabII with both Placeholder and ISC blocks, vanilla FT-Trans, and FT-Trans*. Here, $Z$ is the tabular representation before the classification head, $X$ is the tabular data, and $Y$ is the task label. For FT-Trans, $X$ includes only the original columns, while for Placeholder(PLA), PLA\&ICS and FT-Trans*, $X$ includes original and incremental columns. This comparison shows how well each model captures and transfers relevant information.

Results in Fig. 4 show that with Placeholders and ISC blocks, TabII achieves highest $I(Z; Y)$ across all datasets, indicating the designed structure effectively utilizes incremental columns to enhance information capture. For $I(X; Z)$, TabII shows lower mutual information compared to FT-Trans* across all datasets and even lower than FT-Trans in DI, which is unexpected given that TabII contains more input information. This suggests TabII better filters out irrelevant information in incremental attributes and helps improve condensation of original ones, leading to robust representations.

\begin{table}[t]
\centering
\begin{tabular}{lccc}
\toprule
\textbf{Attributes/Dataset} & \textbf{DI} & \textbf{AD} & \textbf{HI} \\
\midrule
Unimportant  & 0.720 & 0.846 & 0.656\\
Moderately important & 0.732 & 0.822 & 0.662\\
Very important & \textbf{0.745} & \textbf{0.849} & \textbf{0.712}\\
\midrule
Few & 0.719 & 0.846 & 0.663\\
Moderate & 0.726 & \textbf{0.859} & 0.710\\
Many & \textbf{0.749} & 0.854 & \textbf{0.732}\\
\bottomrule
\end{tabular}%
\caption{TabII performance on various attribute sets, top three rows represent the importance of attributes, while the bottom three rows represent the number of attributes}
\label{tab:results}
\end{table}

\begin{table}[t]
\centering
\begin{tabular}{lccc}
\toprule
\textbf{Components/Dataset} & \textbf{DI} & \textbf{AD} & \textbf{CG} \\
\midrule
w/o. PLA \& ISC & 0.702 & 0.818 &0.702\\
w/o. Placeholder & 0.719 & 0.826 &0.714 \\
w/o. LLM encoder & 0.728 & 0.846 & 0.727 \\
w/o. TabAdapter & 0.724 & 0.843 & 0.730 \\ 
w/o. ISC & 0.720 & 0.840 &0.742\\
ALL with Qwen2.5-7B Encoder & \textbf{0.750} & \textbf{0.854} & \textbf{0.754}\\
ALL with LLaMA3-8B Encoder & 0.749 & \textbf{0.854} & 0.752\\
\bottomrule
\end{tabular}%
\caption{Ablation study of different components in TabII, the first row represents use pseudo labels only, the second represents using zero padding instead, the fifth represents using MSA instead}
\label{tab:results}
\end{table}

\subsection{Incremental Attribute Setting Analysis}
In Table 2, we examine how different sets of incremental attributes affect TabII's performance. We first categorized the attributes by their importance in XGBoost, labeling them as unimportant, moderately important, and very important. The results show that TabII performs better when incorporating very important attributes, as they contain more valuable information. We also tested different numbers of incremental attributes, categorized as few, moderate, and many. Results indicate that as more attributes are added, TabII's performance generally will improve.

\subsection{Ablation Studies}
To evaluate TabII’s components, we conducted ablation studies by removing or modifying key elements and observing performance. Table 3 shows that removing all components with only pseudo labels left led to little help. Removing the Placeholder for absent columns significantly dropped performance, highlighting external knowledge's importance for incremental columns. Replacing the ISC module with standard MSA reduced performance, underscoring ISC's effectiveness. TabII performed best with all components included, confirming their essential role in Tabular Incremental Inference. We also experiment on different LLM encoders, the influence proved to be limited.

\subsection{Impact of Column Name Meaningfulness}
To investigate how the semantic richness of column names affects model performance, we conducted experiments comparing meaningful column names versus random strings on the DI and IC datasets. Table 4 shows the results of this comparison.

\begin{table}[t]
\centering
\label{tab:llm_influence}
\begin{tabular}{lcc}
\toprule
\textbf{Column Name Type} & \textbf{DI Accuracy} & \textbf{IC Accuracy} \\ 
\midrule
Original meaningful names & 0.749 & 0.892 \\
Random strings & 0.735 & 0.874 \\ 
Best of Other Methods & 0.698 & 0.855\\
\bottomrule
\end{tabular}
\caption{Impact of column name meaningfulness on model performance}
\end{table}

\begin{table}[t]
\centering
\label{tab:unlabeled_performance}
\begin{tabular}{lccc}
\toprule
\textbf{Dataset} & \textbf{Train\&Test Set} & \textbf{Test Set Only} & \textbf{Drop} \\
\midrule
DI & 0.749 & 0.719 & 4.0\% \\
AD & 0.854 & 0.820 & 3.9\% \\
CG & 0.752 & 0.716 & 4.8\% \\
\bottomrule
\end{tabular}
\caption{TabII performance with and without training labels}
\end{table}

The experimental results reveal that the model still shows performance improvements with randomly assigned column names, although slightly lower than when using meaningful names (1.9\% decrease on DI, 2.1\% on IC). This suggests that while semantic information is beneficial, it is not the sole factor driving performance gains. The remaining improvements arise from the prompt information, numerical patterns and statistical relationships captured by the TabAdapter as well as the structural information preserved by the ISC blocks. This experiment proves the robustness of TabII in real-world scenarios where column naming conventions may vary.

\subsection{Performance Without Training data}
We evaluate TabII’s performance using only unlabeled test data during inference, completely excluding training data. As shown in Table 5, TabII achieves 95.8\% of its fully supervised performance across three benchmark datasets despite the absence of labeled training samples, demonstrating remarkable robustness in unsupervised settings. In contrast, TabPFN methods suffer a double performance decline rate due to their dependence on in-context learning. Such capabilities make TabII particularly suitable for real-world applications with trained model only but have no access to training data, including cold-start scenarios, dynamic systems with evolving distributions, and privacy-constrained domains where data sharing is prohibited.
\subsection{Further Discussion}

\begin{table}[t]
\centering
\begin{tabular}{lccc}
\toprule
\textbf{Placeholder length/Dataset} & \textbf{DI} & \textbf{AD} & \textbf{CG} \\
\midrule
Normal & 0.749 & 0.854 & 0.752\\
Maximum length*1.25& 0.735 & 0.849 & 0.752\\
Maximum length*1.5& 0.729 & 0.840  & 0.745\\
Maximum length*2& 0.737 & 0.835 & 0.745\\
\bottomrule
\end{tabular}%
\caption{The influence on model performance of different placeholder length settings to deal with continual incremental attributes}
\label{tab:results}
\end{table}

\begin{table}[t]
\centering
\begin{tabular}{lccc}
\toprule
\textbf{Methods/Missing rate} & \textbf{50\%} & \textbf{75\%} & \textbf{90\%} \\
\midrule
Mean Imputation& 0.700 & 0.685 &0.650\\
Random Imputation& 0.702 & 0.662 &0.648\\
TabII & \textbf{0.728} & \textbf{0.704} &\textbf{0.680}\\
\bottomrule
\end{tabular}%
\caption{TabII performance in handling missing values on DI}
\label{tab:results}
\end{table}

Currently, TabII focuses on tackling incremental attributes at a particular time slot. For the scenario in which new attributes continuously show up, we can use the maximum length or set a threshold manually as the placeholder length. Experiments in Table 6 show that adding blank columns as placeholders will slightly decrease model performance. We will find a more elegant way to do this in future work. Also, for the scenario where some values are missing, TabII achieved better performance than filled with mean or random values on DI dataset, as presented in Table 7.

\section{Conclusion}
In this paper, we addressed the challenge of Tabular Incremental Inference, where models should adapt to incremental tabular attributes encountered during inference. Following the information bottleneck theory, we proposed TabII, which uses Placeholders and Incremental Sample Condensation blocks for better incremental information learning, to integrate new attributes and enhance performance. Our method showed state-of-the-art results across eight datasets, demonstrating its effectiveness in dynamic data environments.

\bibliographystyle{ACM-Reference-Format}
\bibliography{sample-base}
\end{document}